# Improving FHB Screening in Wheat Breeding Using an Efficient Transformer Model

Babak Azad[1], Ahmed Abdalla[1], Kwanghee Won[2], and Ali Mirzakhani Nafchi[1]

[1] Agronomy, Horticulture, & Plant Science Department, South Dakota State University, Brookings, SD 57007, USA
[2] Electrical Engineering and Computer Science Department, South Dakota State University, Brookings, SD 57007, USA

## Abstract

*Fusarium head blight (FHB) is a devastating disease that causes significant economic losses annually on small grains, such as wheat and barley. Efficiency, accuracy, and timely detection of FHB in the resistance screening are critical for wheat and barley breeding programs. In recent years, various image processing techniques have been developed using supervised machine learning algorithms for the early detection of FHB. The state-of-the-art convolutional neural network-based methods, such as U-Net, employ a series of encoding blocks to create a local representation and a series of decoding blocks to capture the semantic relations. However, these methods are not often capable of long-range modeling dependencies inside the input data, and their ability to model multi-scale objects with significant variations in texture and shape is limited. Vision transformers as alternative architectures with innate global self-attention mechanisms for sequence-to-sequence prediction (used in other fields), due to insufficient low-level details, may also limit localization capabilities. To overcome these limitations, a new Context Bridge is proposed to integrate the local representation capability of the U-Net network in the transformer model. In addition, the standard attention mechanism of the original transformer is replaced with Efficient Self-attention, which is less complicated than other state-of-the-art methods. To train the proposed network, 12,000 wheat images from an FHB-inoculated wheat field at the SDSU research farm in Volga, SD, were captured. In addition to healthy and unhealthy plants, these images encompass various stages of the disease. A team of expert pathologists annotated the images for training and evaluating the developed model. As a result, the effectiveness of the transformer-based method for FHB-disease detection, through extensive experiments across typical tasks for plant image segmentation, is demonstrated.*

## Introduction

As the global population continues to grow, it puts immense pressure on food resources, especially small grain crops such as wheat, barley, and oats, which are vital staples for a large portion of the world's food supply. Among the most significant threats to these crops is Fusarium Head Blight (FHB), a disease that can lead to substantial yield reductions and negatively impact grain quality, animal, and human health [1]. Addressing FHB is essential for ensuring food security and maintaining the quality of grains while protecting animal and human well-being. Traditional methods for detecting and managing FHB can be labor-intensive, time-consuming, and require specialized knowledge, potentially causing contamination and health risks due to excessive chemical use. Artificial Intelligence (AI) and machine vision technologies offer a more efficient and effective solution by automating disease monitoring processes and improving crop productivity and quality. However, accurately diagnosing infected crops and identifying infection levels remain significant challenges [2].

Automatic image semantic segmentation using deep convolutional neural networks (CNNs) has shown promise in precisely segmenting regions displaying disease patterns in input images. Despite their success, CNNs often lose details at deeper layers, prompting the development of U-shaped networks like U-Net [3], which have demonstrated exceptional performance. Although CNN-based models can improve context modeling to some extent, they are inherently limited due to their convolution-based nature. Transformer architectures, known for their success in natural language processing (NLP), have demonstrated the ability to learn long-term features, making them suitable for addressing these limitations [4], [5].

Researchers have explored the impact of Transformer networks on computer vision, with Vision Transformer (ViT) [6] achieving high performance in image recognition tasks compared to state-of-the-art methods. Transformers have proven effective in a wide range of computer vision tasks [7], [8], [9], and their integration with U-Net has been studied to simultaneously benefit from Transformers' global context modeling and CNNs' ability to learn rich local information. However, limitations persist, such as insufficient local context modeling and inadequate integration of multi-scale information generated by hierarchical encoders [28].

In this paper, inspired by the Missformer approach [22] in medical image processing, we propose a powerful and efficient image segmentation transformer model that leverages the long-range dependency capabilities of self-attention for accurate image segmentation. The model is based on the U-shaped architecture, featuring a redesigned Enhanced Transformer Block for improved feature representations. Our model includes an encoder, bridge, decoder, and skip connection, all built upon the enhanced transformer block. The encoder extracts hierarchical features from overlapped image patches, while the bridge models local and global dependencies between different scale features. The decoder generates pixel-wise segmentation predictions with a skip connection. The main contributions of this paper are:

• The proposal of a position-free and hierarchical U-shaped transformer for FHB disease segmentation.
• The redesign of a powerful feed-forward network for better feature discrimination, long-range dependencies, and local context, leading to the development of an Enhanced Transformer Block for robust feature representation.
• The implementation of an Enhanced Transformer Context Bridge based on the Enhanced Transformer Block to capture local and global correlations of hierarchical multi-scale features.
• Outstanding experimental results on Wheat FHB image segmentation datasets, demonstrating the proposed model's effectiveness, superiority, and robustness.

# Literature Review

**CNN-based methods**

Convolutional neural networks (CNNs) have been widely used for various computer vision tasks, including image classification, object detection, and semantic segmentation. In recent years, numerous CNN-based methods have been developed for the early detection and segmentation of plant diseases, including FHB [10], [11], [12]. These methods have achieved significant improvements in terms of accuracy and efficiency over traditional, non-AI-based techniques.
One of the most prominent CNN-based architectures is U-Net, which was introduced by Ronneberger et al. [3]. U-Net features a U-shaped structure consisting of symmetric encoders and decoders with skip connections, allowing the network to capture high-level and low-level information. The U-Net architecture has been successfully applied to various image segmentation tasks, including medical imaging and agricultural applications.

Despite the success of CNN-based methods, they have limitations in modeling long-range dependencies and capturing multi-scale information, which is essential for accurate semantic segmentation. To address these issues, researchers have developed various techniques, such as dilated convolutions [13], atrous spatial pyramid pooling (ASPP) [14], and feature pyramid networks (FPN) [15]. These methods have demonstrated improved performance in capturing contextual information and handling objects of different scales in the input data.

**Transformer-based methods**

The introduction of the Transformer architecture by Vaswani et al. [16] has revolutionized the field of natural language processing (NLP). Transformers employ a self-attention mechanism that enables them to capture long-range dependencies more effectively than traditional CNN-based approaches. Recently, researchers have noticedthe potential of Transformer architectures for computer vision tasks, including image classification, object detection, and semantic segmentation.

The Vision Transformer (ViT) by Dosovitskiy et al. [6] was one of the first attempts to apply Transformer architectures to image recognition tasks. ViT achieved high performance compared to state-of-the-art CNN-based methods by dividing an image into non-overlapping patches and feeding them into a standard Transformer with positional embeddings. This approach demonstrated the potential of Transformer architectures for handling visual tasks.

Subsequent works, such as Pyramid Vision Transformer (PVT) by Wang et al. [17] and Swin Transformer by Liu et al. [18], further explored the application of hierarchical Transformers for computer vision tasks. These methods introduced spatial reduction attention (SRA) and window-based attention, respectively, to reduce computational complexity while maintaining high performance.

Recently, researchers have attempted to combine the strengths of both CNNs and Transformers for semantic segmentation tasks. Uformer by Wang et al. [19], SegFormer by Xie et al. [20], and PVTv2 by Wang et al. [21] embedded convolutional layers between fully connected layers of the feedforward network in Transformer blocks to capture local context and improve feature discrimination. However, these methods still have limitations in handling multi-scale information and capturing long-range dependencies [29].

In this paper, drawing inspiration from the Missformer method [22] in the medical image processing field, we propose a Transformer-based method for FHB disease segmentation, which aims to address the limitations of both CNN-based and Transformer-based approaches by leveraging their respective strengths in capturing local context and modeling long-range dependencies. The proposed method features an Enhanced Transformer Block and an Enhanced Transformer Context Bridge to provide accurate image segmentation results for FHB disease detection.

# Proposed Method

In this section, we first provide an overview of the proposed pipeline and its specific structure, followed by a detailed explanation of the Enhanced Transformer Block, which serves as the foundation of the proposed pipeline. Subsequently, we introduce an Enhanced Transformer Context Bridge that models hierarchical multi-scale information correlations at local and global levels.

**Overall Pipeline**

A visual representation of the proposed model is depicted in Figure 1. The model is inspired by Missformer [22] and involves a hierarchical encoder-decoder architecture incorporating a transformer context bridge module between the encoder and decoder. The input image is divided into four overlapping patches to maintain local continuity. The encoder generates multi-scale features from these patches. Each stage of the hierarchical encoder includes enhanced transformer blocks and patch merging layers. A patch merging layer is employed to produce down-sampling features with the enhanced transformer block, which learns long-range dependencies and local context with minimal computational complexity.

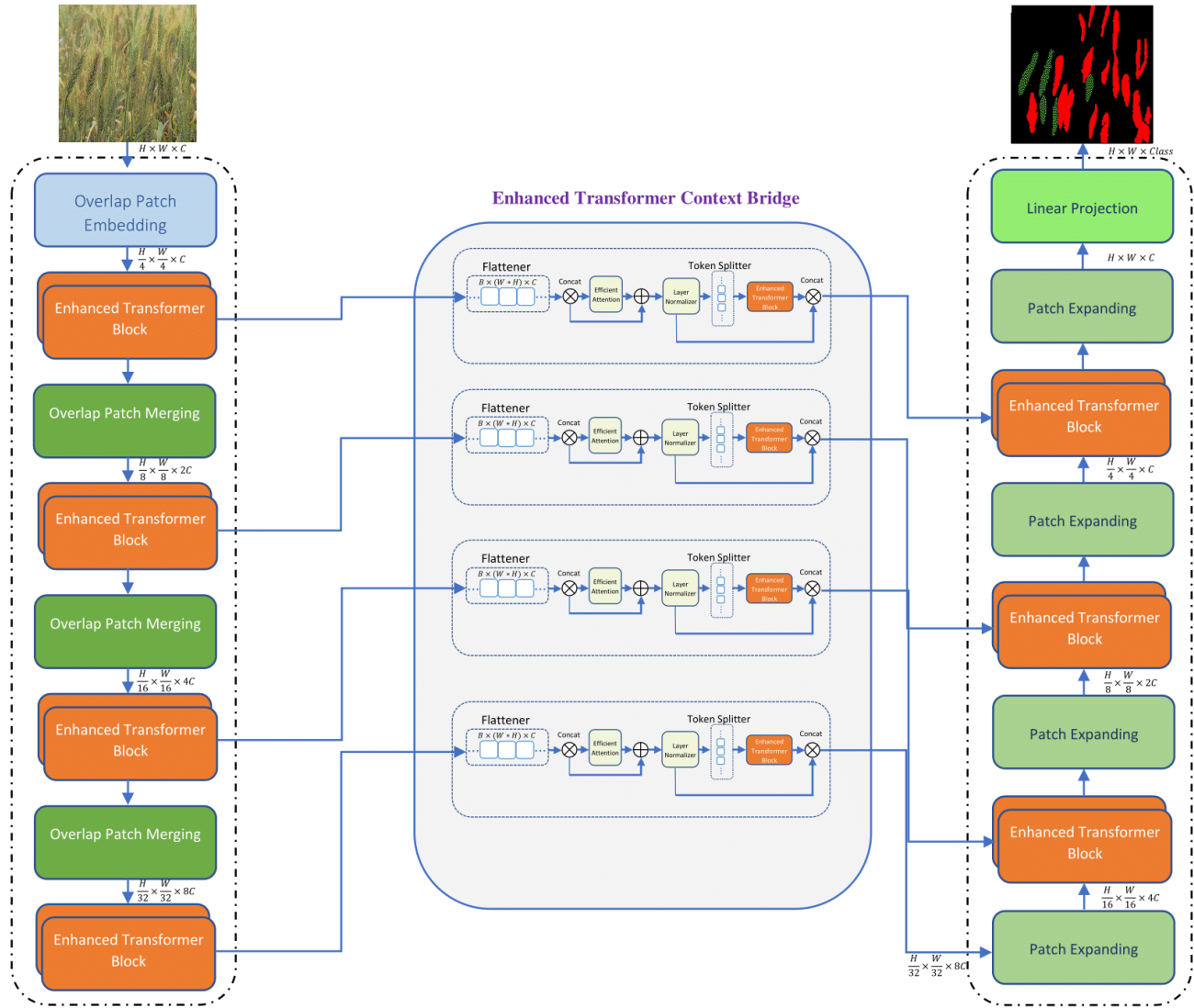

*Figure 1. The structure of the proposed method for FHB disease segmentation. The model utilizes a Transformer module at the network's bottleneck to learn long-range contextual relationships and produces a spatial normalization coefficient for the attention module.*

The model passes the generated multi-scale features through the Enhanced Transformer Context Bridge to capture local and global correlations between different scale features. Flattened spatial dimensions are created and reshaped to maintain consistent channel dimensions, which are then concatenated in a flattened spatial dimension and fed into the enhanced transformer context bridge. Next, we split and restore the data to its original shape to obtain discriminative hierarchical multi-scale features.

The model uses these discriminative features and skip connections as inputs for decoders. Each decoder stage comprises Enhanced Transformer Blocks and a patch-expanding layer [23]. In contrast to the patch merging layer, the patch expanding layer up-samples adjacent features to double the resolution, except for the final stage, which has four times the resolution. Additionally, linear projection generates pixel-wise segmentation predictions.

**Enhanced Transformer Block**

Precise FHB disease segmentation relies on capturing long-range dependencies and local context. Convolutional neural networks represent local features, while transformers are adept at handling long-range dependencies. However, the original transformer block's computational complexity grows quadratically with the feature map resolution, rendering it unsuitable for high-resolution maps. Additionally, transformers face difficulties extracting local context [24], [25]. Although Uformer, SegFormer, and PVTv2 have attempted to overcome this limitation by directly incorporating convolutional layers into feedforward networks, these methods limit feature discrimination while improving performance.

To tackle this challenge, we use a hybrid Transformer Block that combines LayerNorm, Efficient Self-attention, and a local feature representation capturing block, as depicted in Fig. 2.

**Enhanced Transformer Context Bridge**

Fusion of multi-scale information has been demonstrated to be essential for achieving precise semantic segmentation in CNN-based methods [26]. This section explores the fusion of multi-scale features for Transformer-based methods. As depicted in Fig. 1, multiple stage feature maps are generated after input patches are processed by the encoder, maintaining the same settings for patch merging and channel depth across all stages. With multi-level features F1, F2, F3, and F4 produced by the hierarchical encoder, we flatten and reshape them to ensure consistent channel depth between them. Next, we concatenate these features in the flattened spatial dimension and pass the resulting token into the enhanced transformer block to establish long-range dependencies and local context correlations. Once the features have traversed through d enhanced transformer blocks, we divide the tokens, revert them to their original shapes for each stage, and input them into a transformer-based decoder along with a corresponding skip connection for pixel-wise segmentation map prediction. In our approach, the Context Bridge depth is set to 4.

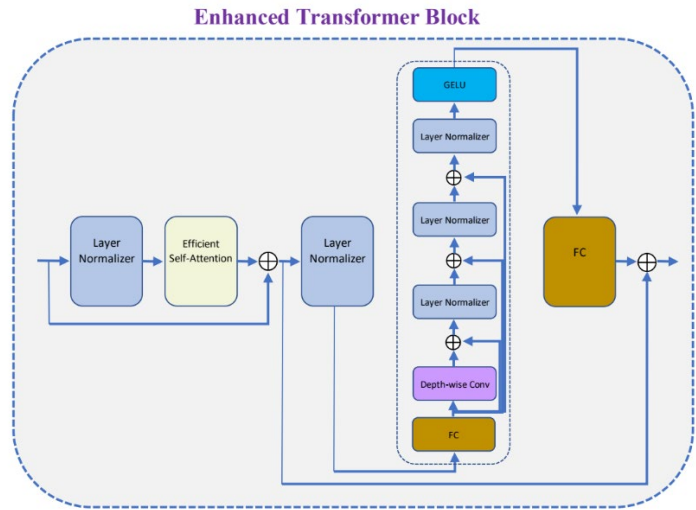

Figure 2: The architecture of the Enhanced Transformer Block.

# Experimental Results

In this section, we present information regarding the dataset, training procedure, the evaluation metrics employed in our experiments, and a comprehensive examination of the experimental outcomes.

**Dataset**

In this study, considering the lack of a comprehensive dataset encompassing all stages of wheat FHB disease, we have collected a large-scale Wheat FHB disease image dataset for the first time. This dataset was introduced in a poster presentation at the 2022 National Fusarium Head Blight Forum in Tampa, Florida, USA [27].

To train our proposed network, 12,000 wheat images were manually captured from an FHB-inoculated wheat field at the SDSU research farm in Volga, SD. These images were taken using a Canon EOS D5 VI camera with an EF 24-105mm f/4L IS II USM lens. A team of expert pathologists annotated the images, which were then utilized for training and evaluating the developed model. Figure 3 displays some samples from our dataset.

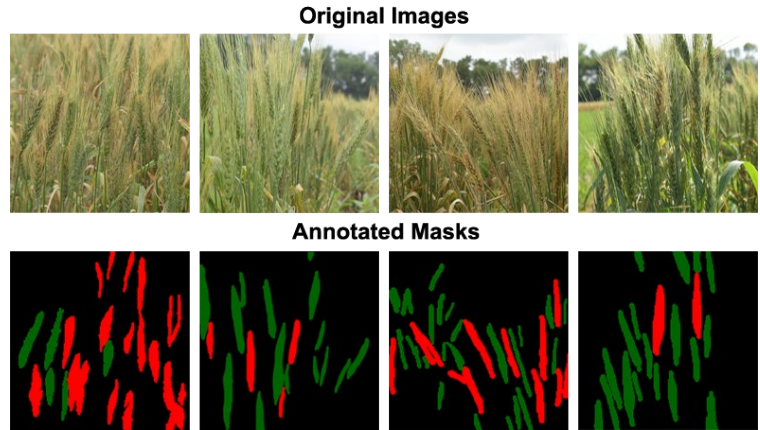

*Figure 3: Sample images from the wheat FHB disease dataset. These images, captured at the SDSU research farm in Volga, SD, demonstrate various stages of FHB disease progression and have been annotated by expert pathologists for training and evaluation purposes. Green annotations denote healthy sections, while red annotations indicate areas affected by the disease.*

For the testing phase, we designed an automatic rotational deep scanner robot to capture images from fields autonomously. As depicted in Figure 4, the adjustable mobile structure, rotational camera platform, and drive motor are crucial design parameters for accurate disease detection. To obtain high-quality close-up shots of wheat spikes, the camera must maintain a high angle toward the subject while preserving a consistent height. Moreover, the lens distance should be carefully calibrated to ensure it remains at an appropriate distance from the rotation center to capture the perfect shot. A counterweight balance is necessary to account for the momentum generated by camera rotation and achieve angular speed equilibrium. Additionally, it is essential that the structure's velocity provides sufficient speed for the camera to capture high-resolution shots.

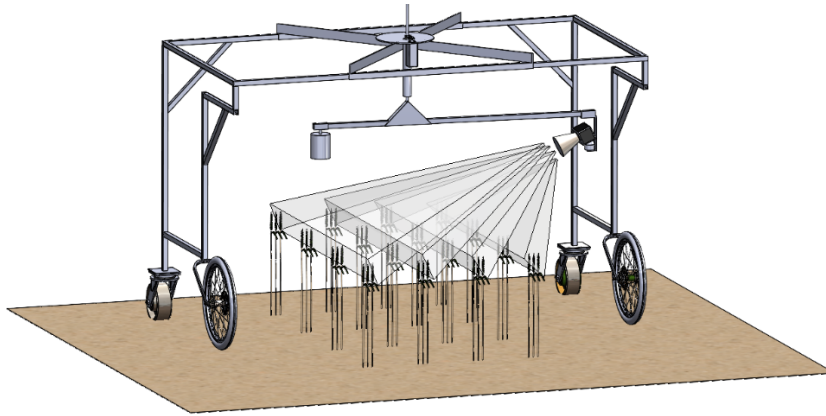

*Figure 4: Schematic of the automatic rotational deep scanner robot designed for capturing test images in wheat fields. The adjustable mobile structure, camera rotational platform, and drive motor are crucial design components to ensure accurate disease detection. The figure highlights key parameters such as camera angle, height, lens distance, counterweight balance, and structure velocity for capturing high-resolution shots.*

### training procedure

Training Procedure: In this study, we implemented the proposed method using Pytorch on a Tesla V100 GPU with 32GB VRAM. We employed an 8-batch batch size without any data augmentation. The models were trained for 100 epochs at a learning rate of $1e-3$ and a decay rate of $1e-4$. The weights of the model were initialized using a standard normal distribution to ensure stability and minimize weight fluctuation. Additionally, the training process was terminated if the validation performance remained unchanged for ten consecutive epochs. The optimization algorithm consistently reduced the loss value on the training and validation datasets throughout the training process, eventually converging to the optimal solution. No signs of instability were observed during the training process.

### Metrics

Metrics: To ensure the validity of our results and draw conclusions about the applicability of our approach, we employed various well-known comparison metrics, including accuracy (AC), sensitivity (SE), specificity (SP), F1-Score, and Jaccard similarity (JS). The terminologies used to describe how metrics are calculated are listed below:

- True-Positive (TP): The predicted label that is correctly identified as a true class.
- False-Positive (FP): The predicted label that is incorrectly predicted as a true class.
- True-Negative (TN): The predicted label that is accurately labeled as a false pixel.
- False-Negative (FN): The predicted label that is mistakenly labeled as a false pixel.

Accuracy: gives the percent of correct predictions,
$$ACC = \frac{TP+TN}{TP+TN+FP+FN} \quad (1)$$

Specificity: determines what percentage of FPs the model correctly identifies,
$$Specificity = \frac{TN}{TN+FP} \quad (2)$$

Sensitivity: determines what percentage of TPs the model correctly identifies,
$$Sensitivity\ (Recall) = \frac{TP}{TP+FN} \quad (3)$$

Balanced F-score (also known as F1 score and F-measure): refers to the weighted average of precision and recall,
$$F1score = \frac{2*TP}{2*TP+FP+FN} \quad (4)$$

Jaccard similarity (also known as a mean intersection over union): Compares members of two sets to determine which members are common and which are distinct between predicted values Ý and real values y,
$$Jaccard\_similarity = y \cap \hat{y} \vee \frac{}{|y|+\hat{y}v - y \cap \hat{y}v} \quad (5)$$

**Quantitative and Visual Results**

In this section, we employed the introduced dataset and metrics to compare the performance of our proposed method with three other models: the basic U-Net [3], Dual-Attention U-Net [30], and ViT model [6]. Our objective was to demonstrate the superiority of our model over these alternatives in FHB disease segmentation. To ensure a fair comparison, we conducted experiments using the same training and testing dataset splits for all models. In our setup, we allocated 70% of the dataset for training, 15% for validation, and 15% for testing.

As shown in Table 1, the results demonstrate that our proposed method outperformed all three state-of-the-art approaches in all metrics. The improvements can be attributed to the effective integration of local context and long-range dependencies enabled by the Enhanced Transformer Block and Enhanced Transformer Context Bridge in our model. This led to more accurate FHB disease segmentation and better disease detection.

*Table 1: Quantitative results comparison of the proposed method with the basic U-Net, Dual-Attention U-Net, and the ViT model in terms of accuracy (AC), sensitivity (SE), specificity (SP), F1-Score, and Jaccard similarity (JS) for FHB disease segmentation. The proposed method demonstrates superior performance across all metrics.*

| Method | F1 | SE | SP | AC | JS |
|---|---|---|---|---|---|
| U-Net [3] | 0.869 | 0.910 | 0.907 | 0.912 | 0.899 |
| Dual-Attention U-Net [30] | 0.924 | 0.913 | 0.984 | 0.970 | 0.970 |
| ViT [6] | 0.947 | 0.936 | 0.988 | 0.981 | 0.979 |
| **Proposed Method** | 0.951 | 0.942 | 0.989 | 0.986 | 0.981 |

The quantitative results demonstrated the superiority of our proposed method in FHB disease segmentation, outperforming the other methods in terms of accuracy, sensitivity, specificity, F1-Score, and Jaccard similarity. This highlights the potential of our model to be employed as a reliable tool for FHB disease detection and management in wheat fields.

Besides the quantitative results, we also provide a visual comparison of the segmentation results obtained using our proposed method and other existing methods, including U-Net, Dual-Attention U-Net, and ViT. Figure 5 displays the segmentation outcomes for each method, allowing for a more intuitive assessment of their performance. The segmentation results reveal that our proposed method generates smooth segmentation outcomes and accurately distinguishes diseased parts from the healthy portions of the plant. This visual representation further emphasizes the effectiveness of our approach in comparison to the other methods.

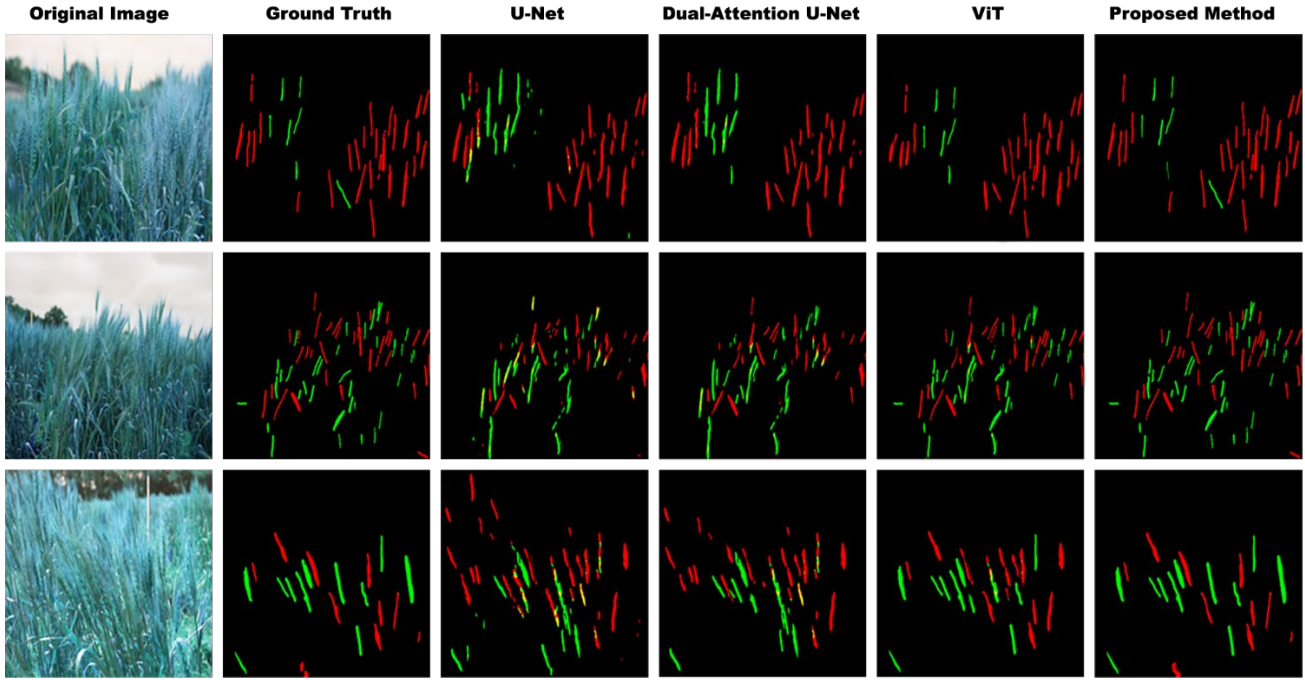

*Figure 5: Visual comparison of segmentation results for the proposed method, U-Net, Dual-Attention U-Net, and ViT. Each row represents a different input image, while the columns show the respective segmentation results obtained by the various methods.*

## Conclusion

This paper presents a novel Transformer-based method for FHB disease segmentation, effectively addressing the limitations of both CNN-based and Transformer-based approaches by leveraging their strengths in capturing local context and modeling long-range dependencies. Our method employs an Enhanced Transformer Block and an Enhanced Transformer Context Bridge to achieve accurate image segmentation results for FHB disease detection. We have also introduced a large-scale Wheat FHB disease image dataset and demonstrated the effectiveness of our proposed method by comparing it to traditional U-Net, Dual-Attention U-Net, and ViT models. The quantitative results reveal that our method outperforms these techniques, highlighting its potential for practical applications in FHB disease detection and management. Future work may involve exploring other Transformer-based architectures and incorporating data augmentation techniques to improve the performance of the proposed method further. In addition, developing an end-to-end system for FHB disease detection, monitoring, and treatment could also be pursued, leveraging the strengths of our method and the presented dataset. Overall, this study contributes significantly to the ongoing efforts to advance the field of FHB disease detection and provides a promising foundation for future research and development in this critical area.

## Acknowledgement

The authors thank their colleagues at SDSU University, Dr. Karl Glover, Dr. Sunish Kumar Sehgal, and Dr. Shaukat Ali, for their helpful feedback and support throughout the research process. Their expertise, insights, and encouragement were invaluable in helping us to complete this work. The authors also want to thank the South Dakota Wheat Commission and the U.S. Department of Agriculture for providing financial support for our research under Agreement 59-0206-2-143, 59-0206-2-153, and FY22-SP-002.